\documentclass{article}

\usepackage{arxiv}

\usepackage[utf8]{inputenc} % allow utf-8 input
\usepackage[T1]{fontenc}    % use 8-bit T1 fonts
\usepackage{hyperref}       % hyperlinks
\usepackage{url}            % simple URL typesetting
\usepackage{booktabs}       % professional-quality tables
\usepackage{amsfonts}       % blackboard math symbols
\usepackage{amsmath}
\usepackage{nicefrac}       % compact symbols for 1/2, etc.
\usepackage{microtype}      % microtypography
\usepackage{siunitx}
\usepackage{graphicx}
\usepackage{subfig}

\author{
	Lichao Chen\\
	University of California, Los Angeles\\
	\texttt{lichao@ucla.edu} \\
	\And
	Sudhir Singh\\
	University of California, Los Angeles\\
	\And
	Thomas Kailath\\
	Stanford University\\
	\texttt{kailath@stanford.edu} \\
	\And
	Vwani Roychowdhury\\
	University of California, Los Angeles\\
	\texttt{vwani@ucla.edu} \\
}

\graphicspath{{img/}{si_img/}{main_img/}}

\title{Brain-Inspired Automated Visual Object Discovery and Detection}

\begin{document}
	\maketitle
	
\begin{abstract}
	In spite of significant recent progress, machine vision systems lag considerably behind their biological counterparts  in  performance, scalability and robustness. A distinctive hallmark of the brain is its ability to automatically discover and model objects, at multi-scale resolutions, from repeated exposures to unlabeled contextual data, and then to be able to robustly detect the learned objects under various  non-ideal circumstances, such as partial occlusion and different view angles. Replication of such capabilities in a machine would require three key ingredients: (i) Access to large-scale perceptual data of the kind that humans experience (ii) flexible representations of objects and (iii) an efficient unsupervised learning algorithm. The Internet fortunately provides unprecedented access to vast amounts of visual data. This paper leverages the availability of such data to develop a scalable  framework for unsupervised learning of \textit{Object Prototypes} -- brain-inspired flexible, scale and shift invariant representations of deformable objects (e.g. humans, motorcycles, cars, airplanes, etc.) comprised of  parts, their different configurations and views, and their spatial relationships. Computationally, the object prototypes are represented as geometric  associative networks using probabilistic constructs such as  Markov Random Fields. We apply our framework to various datasets, and show that our approach is computationally scalable and can construct accurate and operational part-aware object models much more efficiently than  in much of the recent computer vision literature. We also present efficient algorithms for detection and localization in new scenes of objects and their partial views. 
	
\end{abstract}

Visual object classification and recognition is of fundamental importance for (almost) all living animals, and evolution has made the underlying systems highly sophisticated, enabling abstractions and specificity at multiple levels of the perception hierarchy. The design of unsupervised, scalable and accurate Computer Vision (CV) systems, inspired by principles gleaned from biological visual processing systems, has long been a cherished goal of the field. For example, the recent success of the Deep Neural Network (DNN) framework has largely been attributed to its brain-inspired architecture, comprised of  layered and locally-connected  neuron-like computing nodes that mimic the organization of the visual cortex\cite{Bengio2009,Hinton2007,LeCun1989,Arel2010,Bengio2013}. The features that a DNN automatically discovers are considered to be its primary advantage \cite{Bengio2013,Hinton2007}, and it outperforms more conventional classifiers driven by hand-crafted features (such as SIFT and HOG \cite{Lowe1999,Dalal2005}).

While the deep learning framework is undoubtedly a significant achievement, especially in simultaneously learning the visual cues of more than a thousand object categories, it is widely acknowledged that brains are much more efficient and that their operating principles are fundamentally different from those of DNNs or other existing machine learning platforms\cite{Cauwenberghs2013b,Ullman2016b}. Some key limitations of existing DNN-like platforms are: (i) \textit{A predominantly supervised framework}, where one must train them  using large manually labeled training sets and (ii) \textit{Lack of a formal framework for bringing in the higher levels of abstraction necessary for developing a robust perceptual framework}, such as recognizing the persistent identity of an object category (e.g., humans, cars, and animals) that is invariant under different views and under variabilities in their shape and form. To put it another way, these platforms lack a framework for a contextual understanding of scenes where different objects and concepts occur together. On the other hand, biological vision systems (i) are largely \textit{unsupervised learning systems} that can learn highly-flexible models for objects based purely on familiarity and repeated visual exposures in different contexts, (ii) can detect such learned objects at various scales and resolutions, and (iii) are highly computationally efficient.  Therefore, \textit{exploring potential synergies between biological and CV systems remains a topic of considerable ongoing interest.}

In this paper, we consider unsupervised machine learning scenarios -- imitating what humans encounter -- such as the following: An automated probe browsing the Internet encounters a large body of \textit{contextual images}, which we also refer to as \textit{perceptual data}, where a majority of the images contain discernible and high quality instances of objects from an unknown set of categories.  For example, images obtained from videos of real-world scenes  show the same objects  persistently in their natural environments. Similarly, contextual visual browsing based on text tags, can provide such large-scale perceptual data. This is exactly the \textit{perceptual learning world view} that, for example, an infant is faced with, and \textit{this is} what the Internet makes available to computers and machines for the first time. It is worth reiterating that in our unsupervised learning framework, and unlike in supervised training scenarios,  \textit{no labels or bounding boxes of any kind are used to tag these images}. Given such perceptual data, the tasks are (i) \textit{to discover and isolate the underlying object categories} just by processing these unlabeled images (ii) \textit{to build visual models of the discovered categories} and then (iii) \textit{to detect instances of the said objects} in new scenarios, all in a \textit{robust manner} not affected by operations such as scaling, occlusion, and different view points. Humans and many other animals routinely execute these and much more complex visual tasks.

As a step toward developing such an unsupervised contextual learning framework, we first abstract some of the key aspects of biological vision systems.  The immediate \textit{goal is not to emulate} the exact granular feature-generating brain hardware, such as neurons and their layered interconnections, \textit{but to try to computationally capture the basic principles} that have been strongly hypothesized as being used in brains and to integrate them into an end-to-end Computer Vision (CV) framework.

\textbf{Object Prototypes-SUVMs:} The related cognitive science review is outlined in more detail in the Supporting Information (SI) Appendix. We have incorporated only certain specific abstractions of the \textit{object prototype theory of perception} \cite{Logothetis1996} in defining what we shall call a Structural Unsupervised Viewlets Model (SUVM) that has three interacting parts to it.

\textbf{(i)}  \textit{Viewlets:} There is strong evidence for the presence of neurons e.g., those in the inferotemporal cortex, that fire selectively in response to  views of different parts of objects. Neurons in this cortex respond selectively to stimuli from color and texture, and even from complex views such as faces\cite{Logothetis1996}. It is as if the brain breaks up an object into visually distinct, but potentially overlapping, jigsaw pieces of different sizes. Each such view is a building block in our model, and to emphasize that such views are not necessarily distinct functional parts, we refer to them as \textit{viewlets}. Thus for our modeling purposes, viewlets are multi-scale visual cues representative of different appearances of the object under different circumstances, e.g. in the case of humans, different views of the head, or arms in different poses, or a half body view, or just the legs in different poses, and partially or fully covered. As explained in the Methodology section, each viewlet is modeled as a distribution over a feature space, allowing for variations in the appearance of  exemplars belonging to the same category.

\textbf{(ii)} \textit{A set of models} that determine how these viewlets  are geometrically organized to create an entire or a partial image of an object, including
%\vspace*{-2ex}
\begin{enumerate}%[label={(\alph*)},font=\bfseries]
	\item A \textit{Spatial Relationship Network} (SRN):  It has been hypothesized that viewlets that have stable geometrical relationships to each other are indexed in the brain according to their relative spatial locations \cite{Desimone1984}.   Exemplars are recognized as class members if and only if the structural information is close enough to that of the prototype \cite{Biederman1992}. We capture this feature through the SRN, which uses a variation of the \textit{spring network model} \cite{Fischler1973}: This is a graph in which nodes are the viewlets and  edges impose  relative distance and scale/size constraints that the viewlets should satisfy. To encode variations, the edges are represented as springs of varying stiffness and length. Since not all parts directly connect to each other  in a physical object, our model naturally allows for sparsity: it introduces springs only among key viewlet pairs that are needed to maintain the integrity of the whole object. Collectively, the entire object model is then defined by the spring-viewlet ensemble.
	\vspace*{-2ex}
	\item A \textit{Configuration-Independent Parts Clustering} (CIPC) that captures certain \textit{semantic structures} of the object prototype by grouping viewlets with distinct appearances into higher-level concepts of \textit{parts}. For example, the notion of the  ``left arm''  is captured by a set of viewlets corresponding to different configurations of the arm, e.g., hanging down or elbows out. Such viewlets look very different from each other in appearance, and yet can be structurally identified as configurations of the same part since they occupy the same relative position with respect to other body parts such as the torso and the head.
	\vspace*{-2ex}
	\item A \textit{Global Positional Embedding } (GPE) in which each viewlet is assigned its own 2-D location and a scale value. An optimization algorithm computes these viewlet-specific location and scale coordinates so that they yield best fits to the relative location and scale constraints in the Structural Relationship Network (SRN). Global Positional Embedding (GPE) brings out the underlying hierarchical semantic structure of the object, i.e., how the viewlets are organized both spatially, and, hierarchically, in scale. Thus, for example, the GPE would show that an upper half-body viewlet subsumes viewlets that correspond to the head and shoulder regions. This spatial map can be further segmented into clusters of viewlets that define important regions of the object prototype, which in turn could be interpreted as higher-level parts of the object category.
\end{enumerate}
%\vspace*{-2ex}
The semantic structures encoded in the \textit{Configuration-Independent Parts Clustering} and in the \textit{Global Positional Embedding}  play an important role in robust detection. For example, detection of matching viewlets (those whose relative location and scale values match predictions made by the object prototype) corresponding to two different body parts, is a much more robust indicator of the presence of a human, than detecting multiple viewlets corresponding to only a single body part. As demonstrated by our results, the Structural Unsupervised Viewlets model leverages both spatial structure and semantic information to enable high-precision object detection. 

\textbf{A Positive-Only Learning setup for estimating SUVMs:} In the methodology section, we formulate a probabilistic model for an SUVM to allow us to develop reliable estimation and learning algorithms. Then we describe how SUVMs for unknown object categories can be \textit{automatically} learned from \textit{large-scale unlabeled perceptual data} (a large body of contextual images). Our framework mimics the process of \textit{perceptual learning} observed in biological systems, wherein category prototypes are learned via repeated exposures to exemplars, and examining them from different perspectives in multiple contexts\cite{Logothetis1996}. Since for our model building we do not need explicit negative examples, we refer to our setup as \textit{positive-only} learning. Our perceptual framework corresponds to closely-related frameworks in the Computer Vision (CV) literature  ranging from \textit{weakly supervised} to \textit{unsupervised}. For example, in the weakly supervised  setup of \cite{Fergus2003}, positive exemplars belonging to a single category are unlabeled, but negative exemplars (e.g., background or clutter images, and images from categories other than the one being learned) are explicitly provided as part of the training set \cite{Fergus2006}. Our model automatically learns to create its own negative examples, and does not need such external information.  In another instance, the unsupervised set-up of \cite{Sivic2005,Cho2015} use  data that has a perceptual bias over multiple categories. For example, in \cite{Cho2015} automated object category discovery is carried out over contextual data that contain unlabeled exemplars belonging to over 20 different categories; this work, however, does not build stand-alone models for each category that can be used to detect instances for new data (see page 23 of SI Appendix). While our paper reports experimental results involving perceptual data that has exemplars belonging to a single category of interest at a time (\textit{thus, closer to the weakly supervised models in CV}), there is nothing in the framework that precludes it from discovering and modeling tens of categories as long as there are sufficiently many instances of each category in the data set (\textit{thus, closer to  unsupervised models in CV}). These weakly supervised and unsupervised models are distinct from the standard \textit{strongly} supervised learning models, where both positive and negative exemplars are labeled. 

The learning step involves joint estimation of the set of unknown viewlets relevant to the object category, and of the associated  models, i.e., SRN, CIPC, and GPE, that constitute the SUVM. We use \textit{Maximum Likelihood Estimation (MLE)} as our foundational framework, and incorporate \textit{sparsity constraints and convex relaxations} to ensure computational tractability. This leads to an intuitive but mathematically rigorous and computationally simple learning framework.\\
\textbf{Object detection and localization:} Given a learned SUVM we turn to the task of detecting  instances of objects in a new image. Again, we abstract how brains are theorized to detect objects as belonging to a category: \textit{by the occurrence of a sufficient number of compatible viewlets that are spatially co-situated as predicted by the SUVM}.  While we follow an MLE framework of locating a portion of the image that has a high likelihood of being generated by the SUVM, we avoid exponential search complexity (usually, associated with the exhaustive combinatorial search required in MLE), by intuitive but careful pruning of the search space based on the structure of the SUVM. This enables a \textit{linear-time} detection and localization algorithm that can locate multiple instances of the object, unaffected by  scale and to the presence of occlusions in the images. The mathematical details are provided in the methodology section. 

\begin{figure*}
	\centering
\includegraphics[width=6.7in]{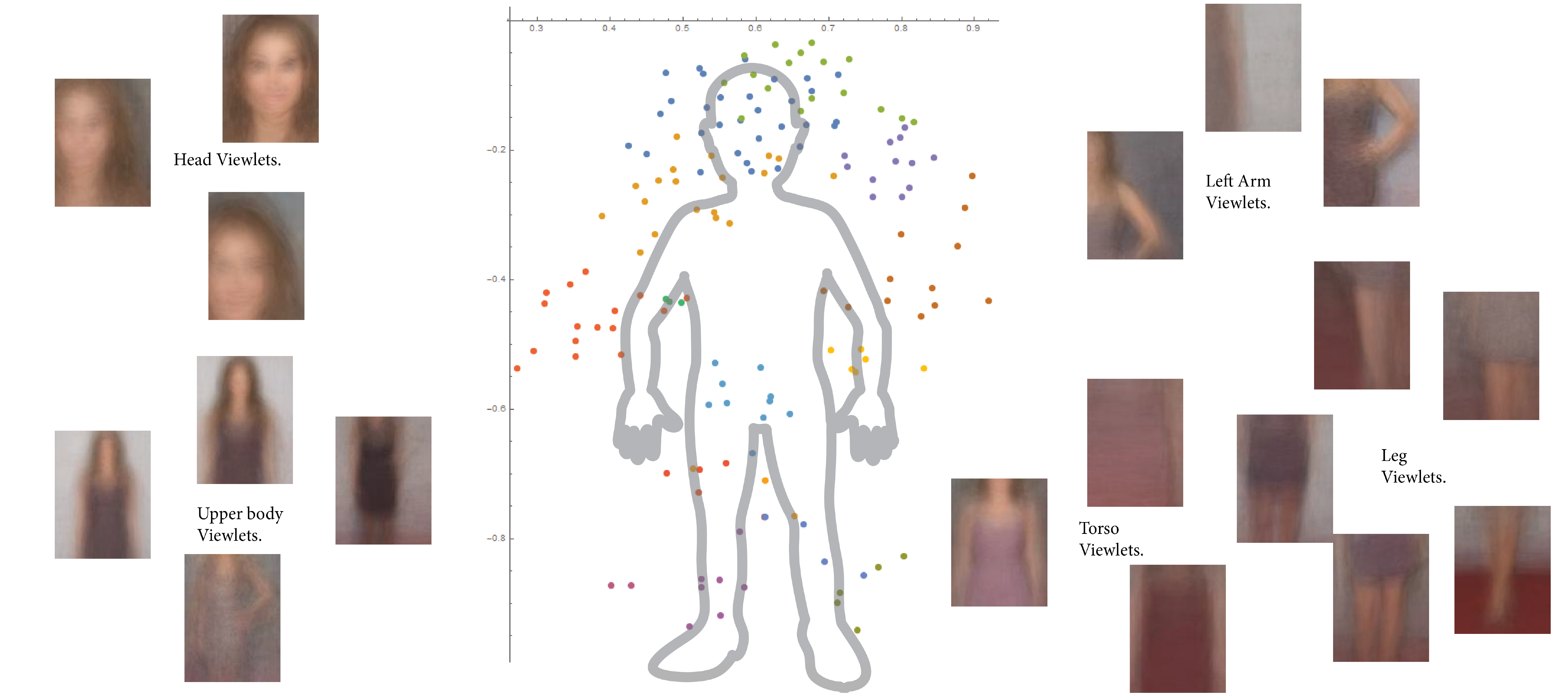}
	\caption{The colored dots in the figure show estimated average $(X,Y)$ coordinates of the centers (as determined by the Global Positional Embedding (GPE) algorithm) of some of the viewlets in our human SUVM. Recall that each viewlet is a distribution over a set of example views/patches that have similar  appearances. The patches belonging to the same viewlet are averaged at the pixel level to create a representative visualization patch. Viewlets automatically clustered as comprising a `part' are given the same color code. For example, we automatically group viewlets corresponding to different views of the left arm  as belonging to the same `part' cluster, which thus can be tagged as the Left Arm part. Three such viewlets corresponding to the arm straight down and with elbow out are shown here. Similarly, multiple viewlets corresponding to different views of head/face, legs, and torso are also shown.  }
	\label{fig:introduction:gps}
\end{figure*}

\textbf{Relation to previous work:} There is a long history of efforts, similar in spirit to ours, aimed at building Computer Vision frameworks that  develop parts-aware object prototypes\cite{Fergus2003,Felzenszwalb2008,Bourdev2009,Chen2014,Chang2011,Cho2015,Sivic2005,Ramanan2013,Fischler1973} (see for example, Geman et. al. \cite{Geman2015} for a review). More details on these models are given in the Results, Discussion and SI Appendix (page 23) sections. Our conclusion is that the goal of learning persistent and flexible object models, especially when the object is deformable and comprised of multiple configurable parts, in an \textit{automated and unsupervised manner} is largely unsolved. Most parts-aware approaches require strongly supervised training\cite{Felzenszwalb2008,Felzenszwalb2010,Chang2011,Ramanan2013}. Moreover, to compensate for computational scalability challenges, they use limited object models such as (i) using only a few parts in describing the object, and (ii) further restricting the kinds of relationship patterns between parts so that they form networks such as trees and stars. \textit{Almost all previous attempts at unsupervised (e.g., \cite{Cho2015} that can be interpreted as using a star-network spring model) and weakly supervised (e.g., \cite{Fergus2006}) frameworks require datasets where exemplars have very similar views}. For example, exemplars must have arms in the same position relative to the body, e.g., see Fig.~\ref{fig:introduction:gps} on page~\pageref{fig:introduction:gps}, where we illustrate this point. On the other hand, our SUVM framework is flexible, computationally scalable and allows for models comprising hundreds of viewlets per object, all embedded in a flexible spatial model and a semantic structure that is independent of visual appearances. From a purely modeling perspective, SUVM can be considered as combining ideas from both the supervised approaches, e.g., Poselets \cite{Bourdev2009} (which uses an ensemble of templates with embedded exemplars (with tags at key points) to define an object category), and the flexible mixtures of parts model for human pose detection and estimation~\cite{Ramanan2013} (which can be interpreted as grouping viewlets into an ensemble of tree networks, instead of a single SRN) and the probabilistic parts-constellation models used in the weakly-supervised approach of \cite{Fergus2003,Fergus2006} and the unsupervised approaches in \cite{Cho2015,Sivic2005}.

%------------------------------------------------------------------------
\section{Results}
\label{sec:results}

%------------------------------------------------------------------------
\subsection{Data Description}
\label{subsec:result:data}
We experimented with two datasets: \\ \textbf{(i)} The CalTech-4 data set, comprising Faces (435 images), Motorbikes (800 images), Airplanes (800 images) and Cars (800 images). A primary motivation for selecting this data set is that there are existing results, using  earlier \textit{weakly} supervised and unsupervised parts-aware formalisms\cite{Fergus2003,Fergus2006}, which we can use to evaluate our framework (See Tables 1 and 2). The other motivation is that it enables us to perform a number of experiments to explore the limits of our SUVM framework. It allows us, for example, to evaluate how our learning framework performs with small-size data (e.g., only 218 images are available to learn models for a face; the other 217 being used for testing). The data does not always contain enough exemplars to capture many of the variations in shape and orientations of the object instances. We also ask questions such as: If the machine's world view is limited only to one object category, how would it interpret the rest of the world. For example, would a face SUVM  (\textit{learned only from face related perceptual data and with no exposure to images from other categories}) ``see" faces everywhere when shown images of motorbikes or cars? Furthermore, it allows us to \textit{illustrate another brain-like activity}: the ability to improve detection/localization performance by jointly using multiple object prototype models (learned from different perceptual datasets). Table~\ref{tab:res_vshaar} illustrates how the face SUVM learned from the CalTech-4 dataset can be used in conjunction with the full-body human SUVM to obtain  face detectors with higher precision. 
\\
\textbf{(ii)} A celebrity dataset, which we created by crawling \num{12047} high quality images from the web. This is a good example of the types of perceptual data the Internet can provide, and  is ideally suited to our perceptual framework: the images are from natural settings with diverse backgrounds and resolutions, and often have multiple instances of individuals (the unknown category to be modeled) in the same image, who  display a wide diversity in clothings and body gestures.   For evaluation purposes only, we manually annotated the whole data set with precise main body part (such as the head and the torso) locations; \textit{this information was not used in the learning process}.  There are no unsupervised approaches to extract object prototypes from such data sets, and hence, we compare our detection and localization performance with those of supervised frameworks. See the detailed discussion on \textit{Torso Detection} on page \pageref{subsec:torso-detection}. 
%------------------------------------------------------------------------
\subsection{SUVM Learning and  Visualizations}
\label{subsec:result:learn_n_detect}

\iffalse Most of the details are in the SI and here we focus only on the main results. \fi  
For the celebrity dataset,  we used \num{9638} images as  learning set, and the rest of the images as  test set. As a first step, we learn a set of visual words, which we shall call  a \textbf{visual dictionary},  that captures the repeated visual patterns in the learning set (see discussions in the Methodology section on Page \pageref{subsec:methodology:learn}).  To build the visual dictionary, we first randomly sample \num{239856} image patches (each of size $128 \times 96$ pixels) from the dataset using a \textit{scale pyramid}: we successively scale down each image by a constant multiplicative factor or scale to create a layered `pyramid'. Then we select  fixed-size windows, located at random locations in each layer, to crop image patches. These patches are then represented in the form of dimension-reduced Histogram of Oriented Gradient (HOG) descriptors \cite{Dalal2005}. These descriptor vectors are then grouped into $k$ clusters using the k-Means algorithm. After comparing results of k-Means clustering for different $k$'s, we settled on $k= 1006$, and the corresponding clusters formed our visual dictionary. 

We next followed our SUVM learning steps as described in Section~\ref{subsec:methodology:learn}, and derived a sparse Spatial Relation Network (SRN), containing $566$ viewlets. Thus, while the visual dictionary contains 1006 visual words, only a subset of them are viewlets in the human model, and the rest describe background scenes. These viewlets are then visualized as a weighted average of all the constituent image patches and are shown in the SI Appendix; examples of a few select viewlets and their constituent image patches are also shown in Fig.~\ref{fig:introduction:gps}. As one can determine via visual inspection, each viewlet corresponds to a meaningful human body part. We then (i) constructed a Configuration-Independent Parts Clustering (CIPC) network and \textit{automatically found 18 distinct parts} (see Fig. 8 in the SI Appendix);  and (2) computed a Global Positional Embedding (GPE) of the viewlets. Fig~\ref{fig:introduction:gps} illustrates some of the salient aspects of our human SUV model.

The same steps are executed for the CalTech-4 data sets, with one twist: (i) \textit{Separate Dictionaries}: The visual dictionary for each category is constructed from its category-specific images, and then an SUVM is derived for each category from its own image set (ii) \textit{Shared Dictionary}: A common visual dictionary is derived from all the learning sets, and then four separate SUVMs are learned by processing their respective category-specific images. The resulting viewlets and their automated groupings as parts are illustrated in the SI Appendix.

\subsection{Evaluation of SUVMs via Detection and Localization Tasks}
\begin{figure}
	\centering
	\subfloat[Face viewlets on a Motorcycle image.]{
\includegraphics[width=1.41in]{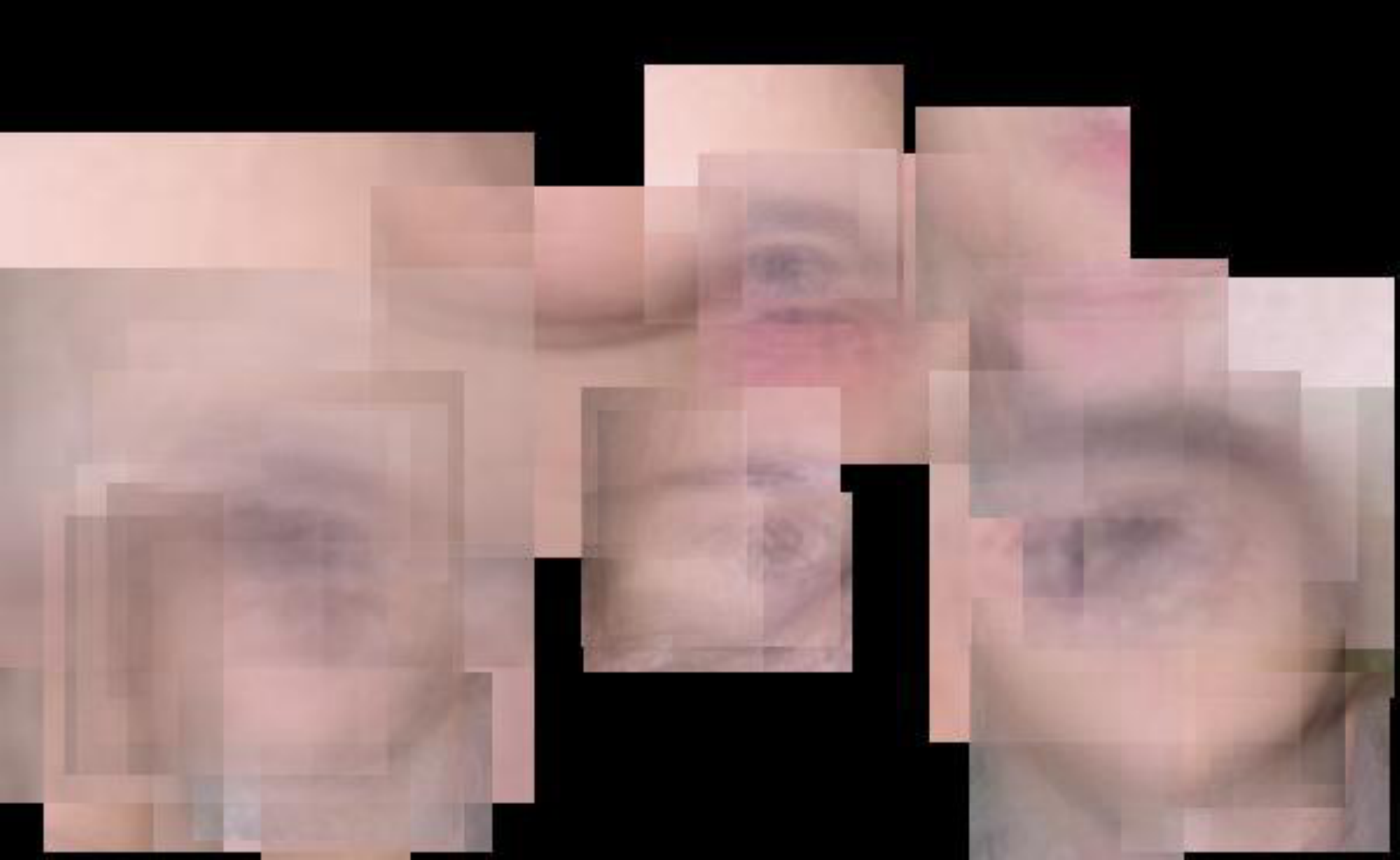}}
	\hfil
	\subfloat[Motorcycle viewlets on the same image.]{
\includegraphics[width=1.41in]{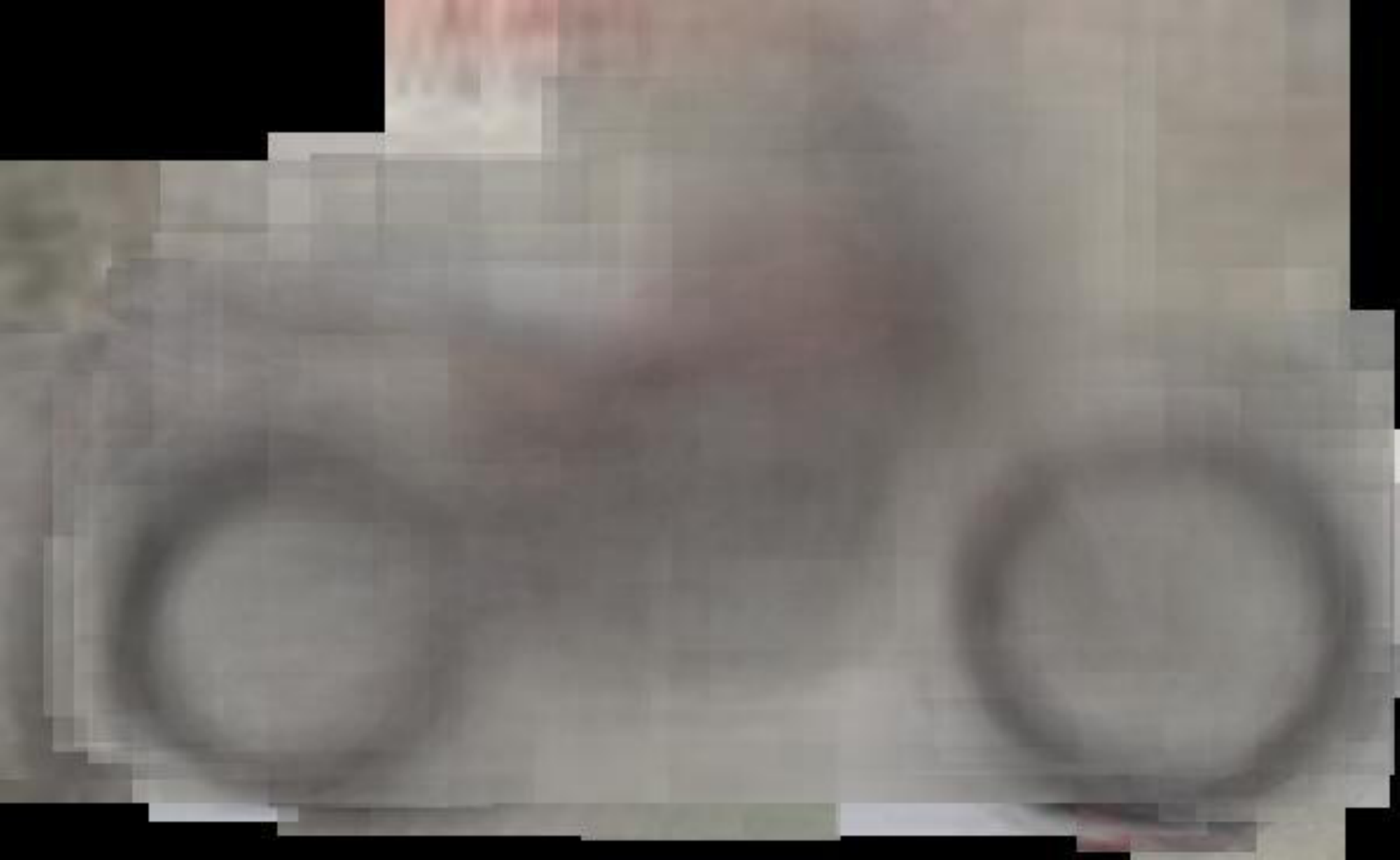} }
	\caption{(a) A motorcycle image viewed through the lens of a face model. While it sees face viewlets everywhere, it does not detect any faces because the viewlets do not match structurally. (b) The same image viewed through a motorcycle model. }
	\label{fig:caltech:face_on_motorbike}
\end{figure}

%\vspace*{2ex}
\textbf{(i)} \textbf{The results on the CalTech-4 data}  set as summarized in Tables~\ref{tab:results:shared} and~\ref{tab:results:separate} show that the SUVM framework significantly outperforms the only other comparable unsupervised/weakly-supervised framework in the literature \cite{Fergus2003,Fergus2006}. \iffalse Recall that the models in \cite{Fergus2003,Fergus2006} do not handle different views or parts in different configurations, and hence cannot be used for processing the much more complex celebrity dataset.\fi As noted earlier, this data set allows us to explore the workings of SUVMs in more detail: For example,   Table~\ref{tab:results:separate} shows that  \textit{a Face SUVM, learned solely from face images, makes no errors} (i.e., does not give false positives of detecting faces) \textit{when fed with images from the rest of the categories}. In comparison, the Face model in \cite{Fergus2003} detected faces in 50\% of the test Motorbike images. Fig.~\ref{fig:caltech:face_on_motorbike}(a) shows an instance of what happens when a motorbike image is viewed through the lens of a face SUVM: \textit{each  patch in the motorcycle image has to be assigned to one of the visual words derived from the face-only data set}, and consequently, individual face viewlets are detected all over the image. However, they do not have the collective structural integrity to be detected as instances of \textit{face}. The Airplane and Car categories are not as discriminative when their visual worlds are based on such separate dictionaries.  As shown in the bottom half of Table~\ref{tab:results:separate}, however, performance improved considerably across three categories (face, motorcycle, and car) when we used a  shared dictionary (i.e., we used images from all the categories to learn a shared set of visual words) but still learned individual SUVMs solely based on their category specific images, and no negative examples were used. The airplane model is the worst performer: not enough examples are present in the learning set (comprised of only 400 images) to learn the different shapes and orientations of airplanes (for example, there are several images with planes pointing in opposite directions) in the data. \iffalse As presented in the SI, however, our platform did learn an accurate object prototype for airplanes with an orientation that matches those of the majority of the exemplars in the data.\fi Finally, by combining the four models together using an SVM (Support Vector Machine), one gets almost perfect detection results across all four categories as shown in Table~\ref{tab:results:shared}.

\iffalse 
The results in Table~\ref{tab:results:separate} are to be contrasted with those in \cite{Fergus2003}, where  background/clutter images were used as negatives, and yet 50\% of the  motorbike images are detected as having faces. The performance was improved in their later work \cite{Fergus2006} by using image patches from other categories as negatives when learning models for a particular category (see Table~\ref{tab:results:shared}); even then a Support Vector Machine (SVM)  classifier was used to combine the four models, and the performance of individual models when tested on images belonging to other categories was not reported. 
 
 For the Caltech 101 dataset, we compare detection results across the four categories for both shared and separate dictionary cases, and the results are summarized in Tables~\ref{tab:results:separate}, and \ref{tab:results:shared}. 
\fi 

\textbf{(ii)} \textbf{Precision and Coverage/Recall Performance for various Detection Tasks and Comparison with Supervised Methods:} Recall that during our object prototype learning process, we automatically break up the given perceptual visual data into a visual dictionary that is comprised of two sets: Visual words that are part of an object prototype, which we refer to as \textit{viewlets}, and the rest that represent visual cues of the background scenes.  \iffalse Since we do not bring in any negative exemplars, the non-object viewlets are the sole .  We have used local Histogram of Oriented Gradients (HOG) features for representing image patches, and then implemented the K-means clustering algorithm for determining the visual words that constitute the dictionary for the model.\fi Since we do not bring in any negative exemplars, \textit{everything the model sees is interpreted only in terms of this dictionary}, and hence, the quality and diversity of the perceptual data plays a very important role in how the model performs when it sees a new image. During detection, the first step is to decompose the given image into patches at multiple scales  using a sliding window and a scale pyramid, and then to assign to each patch the likelihood of it being a particular word from the dictionary that was already learned. We have used the  k-Nearest-Neighbor  classifier for this classification task. 

\textit{The limitations of the kNN classifier are well known} and that is why in most computer vision applications, considerable effort is expended to train much more powerful classifiers using additional negative exemplars that represent other object categories and background scenes.  \textit{It is, however, instructive to note how well an SUVM does}, even with very weak classifiers for viewlets, and without the benefits of training with negative exemplars. In the Discussion section we point out how one can  incorporate negative exemplars to improve performance. 

\begin{table}
	\captionsetup{labelfont=bf,textfont=normalfont}
	\caption{Confusion matrices for the CalTech-4 dataset with   \textbf{one multi-category classifier}: The \textbf{table entry}$(i,j)$ is the percentage of query images belonging to Category$(i)$ that are classified as belonging to Category$(j)$.}
	\label{tab:results:shared}
	\centering	
	\begin{tabular}{l|cccc}
		\toprule
		{}                 & \multicolumn{4}{c}{Classifier $\longrightarrow$ (SUVM + SVM)$\mathbf{/}$(Fergus et al. \cite{Fergus2006})} \\ \cline{2-5}
		Query $\downarrow$ &         (F)ace          &      (M)otorbike       &       (A)irplane       &             (C)ar              \\ \midrule
		F                  & 0.982$\mathbf{/}$0.862  & 0.000$\mathbf{/}$0.073 & 0.018$\mathbf{/}$0.028 &     0.000$\mathbf{/}$0.014     \\
		M                  & 0.000$\mathbf{/}$0.000  & 0.990$\mathbf{/}$0.977 & 0.010$\mathbf{/}$0.013 &     0.000$\mathbf{/}$0.000     \\
		A                  & 0.005/$\mathbf{/}$0.003 & 0.013$\mathbf{/}$0.042 & 0.967$\mathbf{/}$0.888 &    0.015/$\mathbf{/}$0.060     \\
		C                  & 0.000$\mathbf{/}$0.008  & 0.000$\mathbf{/}$0.092 & 0.020$\mathbf{/}$0.197 &     0.980$\mathbf{/}$0.670     \\ \bottomrule
	\end{tabular}
	\caption*{\hspace*{0.05cm} For SUVMs, a  visual dictionary is learned from all the images (i.e. a \textbf{shared visual dictionary} is created), but each model is learned only from its category-specific images. A single 4-class SVM (Support Vector Machine) classifier is  built by combining the outputs of all the four models as was done in \cite{Fergus2006}.}
\end{table}

\begin{table}[h]
	\captionsetup{labelfont=bf,textfont=normalfont}
	\caption{Confusion matrices for the CalTech-4 dataset based on \textbf{four separate category models}. Each category model$(j)$ only outputs whether a query image contains an exemplar of category($j$). The \textbf{table entry}$(i,j)$ is the percentage of query images belonging to Category$(i)$ that are detected to contain an instance of  Category$(j)$ (by using a category model$(j)$). The top table corresponds to the case where SUVMs are created using \textbf{separate dictionaries}, whereas the bottom table corresponds to the \textbf{shared dictionary} case.}
	\label{tab:results:separate}
	\centering
	\begin{tabular}{l|cccc}
		\toprule
		{}                       & \multicolumn{4}{c}{Models $\longrightarrow$ SUVM(Separate)$\mathbf{/}$(Fergus et al. \cite{Fergus2003})} \\ \cline{2-5}
		Query Image $\downarrow$ &           F            &           M           &            A            &               C               \\ \midrule
		(F)ace                   & 0.980$\mathbf{/}$0.964 & 0.069$\mathbf{/}$0.33 &  0.215$\mathbf{/}$0.32  &      0.252$\mathbf{/}$-       \\
		(M)otorbike              & 0.000$\mathbf{/}$0.50  & 0.95$\mathbf{/}$0.925 &  0.370$\mathbf{/}$0.51  &      0.237$\mathbf{/}$-       \\
		(A)irplane               & 0.000$\mathbf{/}$0.63  & 0.007$\mathbf{/}$0.64 & 0.665/$\mathbf{/}$0.902 &      0.025$\mathbf{/}$-       \\
		(C)ar                    &  0.000/$\mathbf{/}$-   &  0.000$\mathbf{/}$-   &   0.002$\mathbf{/}$-    &      0.600$\mathbf{/}$-       \\ \bottomrule
	\end{tabular}
	\begin{tabular}{l|cccc}
		\toprule
		{}                       & \multicolumn{4}{c}{Models$\longrightarrow$ SUVM (Shared) } \\ \cline{2-5}
		Query Image $\downarrow$ &    F     &    M     &    A     &             C             \\ \midrule
		(F)ace                   & 0.972477 & 0.087156 & 0.674312 &         0.073394          \\
		(M)otorbike              & 0.007500 & 0.960000 & 0.675000 &         0.140000          \\
		(A)irplane               & 0.000000 & 0.002500 & 0.745000 &         0.117500          \\
		(C)ar                    & 0.000000 & 0.000000 & 0.167500 &         0.970000          \\ \bottomrule
	\end{tabular}
	\caption*{\hspace*{0.5cm}\iffalse For SUVMs, the visual dictionary and  model for each category are learned only from its category-specific image set. This is a much stricter test of our models, and allows us to explore issues such as the following: If a machine has been exposed to \textbf{only}  human faces, then how confused would it be by a motorbike image.\fi  Column 1 in the \textbf{top table}, for example, shows that the  face SUVM returned no false-positives (FPs) when tested on non-face images; the face model in \cite{Fergus2003}, on the other hand, returned 50\% FPs on Motorbike images. Similarly, the FP rate on face images for the Motorbike model is $6.9\%$ for SUVM vs $33\%$ in \cite{Fergus2003}. The \textbf{bottom table} shows that the Face, Motorbike and Car models do extremely well, even without a separate multi-class classifier. }
\end{table}

\begin{table}[h]
	\captionsetup{labelfont=bf,textfont=normalfont}
	\caption{\textbf{Face Detection:} SUVM vs the Viola-Jones algorithm (VJA) \cite{Viola2001}. \textbf{Coverage/Recall} is the ratio of (\#True-Positives) and (Actual \#Positives in the labeled test data). \textbf{Precision}  is the ratio of (\#True-Positives) and (\#True-Positives + \#False-Positives). }
	\label{tab:res_vshaar}
	\centering
	\begin{tabular}{l | c |c c | c c }
		\toprule
		                & {VJA}  &         \multicolumn{4}{c}{SUVM}         \\ \cline{3-6}
		                &        & H-1    & H-2    & HF-1   & HF-2          \\ \midrule
		True Positive   & 3072   & 2965   & 3048   & 2959   & 3047          \\
		False Positive  & 972    & 54     & 301    & 31     & 183           \\
		Coverage/Recall & 92.9\% & 89.7\% & 92.2\% & 89.5\% & \mbox{92.2}\% \\
		Precision       & 76.0\% & 98.2\% & 91.0\% & 99\%   & 94.3\%        \\ \bottomrule
	\end{tabular}
	\caption*{\hspace*{0.5cm} For a description of the OpenCV implementation used for VJA and ROC plots see page 19 (column 1) of SI Appendix. The columns labeled \textbf{H-1} and \textbf{H-2} represent face detection results for two different settings of parameters, when our human SUVM is used for prediction. A subset of these predicted face patches with high-enough resolution (e.g., those with heights greater than 150 pixels) are then filtered through the Face SUVM, derived from the CalTech-4 dataset and those that do not pass are rejected. The respective results after this filtering are shown in columns  \textbf{HF-1} and \textbf{HF-2}. The human SUVM provides much higher precision while matching the coverage of the well-known algorithm.
	}
\end{table} 

\textbf{(a) Head/face detection:} We used a subset of the viewlets in our Human SUVM, and mapped their detected locations to where the head would be to create a face detector. As summarized in Table~\ref{tab:res_vshaar}, \textit{our results turned out be vastly superior to those obtained by  the  Viola-Jones algorithm \cite{Viola2001}}, which was developed via extensive manual training over multiple years, and was the  face-detection algorithm of choice until the recent development of superior face detectors, made possible by even more extensive training and deep learning. \iffalse \cite{Farfade2015}.  Our precision is  98\% vs 76\% offered by \cite{Viola2001}, while  our recall is only slightly lower: 90\% vs 92.9\%.  The lower recall is due to  the weak kNN classifier  missing instances of many object-related viewlets. \fi  The superior performance is clearly because of the structure embedded in our human SUVM: Viewlets corresponding to other body parts can locate the position of the face, even when our face-only viewlet detectors are weak. As explained in Table~\ref{tab:res_vshaar} \textbf{this experiment allowed us to combine two different SUVMs}: the precision of our face detector (based on our human SUVM) can be significantly increased, without compromising recall performance, by further examining the predicted faces via the face model obtained from the CalTech-4 dataset.  \textit{This emulates how human vision tends to work:} first impressions of objects, based on outlines, are further refined by focusing in on the details.

%\vspace*{1ex}
\textbf{(b) Torso detection:}\label{subsec:torso-detection} Torsos are much harder to localize, due to lack of distinctive features and the large variety introduced by dress patterns. These considerations make a rigid-template-based torso detector impractical. However, part-based models can detect torsos by mapping other detected parts to where the torso would be. \iffalse Thus, torso detection has become a great benchmark for deformable part-based approaches.\fi We compare our model with two other notable part-based approaches, namely Poselets\cite{Bourdev2009} and the Deformable Parts Model (DPM)\cite{Felzenszwalb2008} and the results are summarized in Table~\ref{tab:results:torso}. As shown in Table~\ref{tab:results:torso} we again outperform these strongly-supervised models. Note that the ingenious Poselets approach is based on processing manually tagged data to generate hundreds of distinctive templates (and associated classifiers obtained through extensive supervised training) that have exemplars with various views/poses and scales embedded in them. Such templates, however, do not constitute detailed object prototypes of the kind illustrated in Fig.~\ref{fig:introduction:gps}.\iffalse We optimized the performance of the Poselets model and obtained precision that matched ours, while providing a slightly higher recall as reported in Table 3. \fi The slightly lower recall rate in our model is to be expected as the weak kNN classifer misses many of the viewlets that are otherwise present in an image. As further elaborated in the Discussion section (see the part on \textit{Dealing with low resolution images and Integrating Supervised Learning} on Page~\pageref{subsec:suplearn}), we can also train classifiers for each visual word in our model to improve our performance. \iffalse For detecting each visual word (computed in an automated manner), one can train a Support Vector Machine (SVM) using samples from other viewlets and additional background image patches as negative exemplars.  This will not be necessary as more object prototypes are learned and the ensemble of SUVMs ``see" a lot of different scenarios.\fi The point of this paper is to show how well the SUVM does even with a limited and self-contained world view.  \iffalse Our SUVM, on the other hand,  creates  detailed object prototypes that can be used for both purposes: (a) to effectively detect object instances (without supervised training) in images with sufficient resolution, and (b) as discussed later, to create  templates and associated classifiers (via discriminative learning) to handle low-resolution images.\fi 

\begin{table}
	\captionsetup{labelfont=bf,textfont=normalfont}
	\caption{\textbf{Torso detection:} SUVM vs Parts-Aware \underline{Supervised} Approaches.  }
	\label{tab:results:torso}
	\centering \vspace*{-1.5ex}
	\begin{tabular}{l | r ccc}
		\toprule
		{}                  &                           \multicolumn{4}{c}{Approaches}                           \\ \cline{2-5}
		                    & DPM\cite{Felzenszwalb2008} & Poselets\cite{Bourdev2009} &  SUVM  & SUVM (Stricter) \\ \midrule
		True Positive (TP)  & 1239                       & 3115                       &  2935  &      2838       \\
		False Positive (FP) & 5263                       & 1678                       &  277   &       52        \\
		Coverage/Recall     & 38.3\%                     & 96.3\%                     & 90.7\% &     87.7\%      \\
		Precision           & 19.1\%                     & 65.0\%                     & 91.4\% &     98.2\%      \\ \bottomrule
	\end{tabular}\vspace*{-1.5ex}
	\caption*{For details of DPM and Poselets see page 20 (column 2) of  SI Appendix. SUVM outperforms all the models in precision (for Poselets we used the  recommended threshold value of $3.6$\cite{Bourdev2009}), while providing a solid recall performance. As discussed in the \textit{Torso Detection} Section on Page~\pageref{subsec:torso-detection}, the recall performance of an SUVM can be improved by replacing the kNN classifiers it uses with superior supervised classifiers, and introducing negative examples.  We have intentionally persisted with the weak kNN classifiers to emphasize the power that an SUVM derives from its structure and its hundreds of viewlets.}
\end{table}

\begin{figure}
	\centering
%	\subfloat[Celebrities close to each other.]{
%\subfloat[]{
%		\includegraphics[width=1.31in]{img/fullbody/origin/1522888037783143532_censored}%
		\includegraphics[width=1.31in]{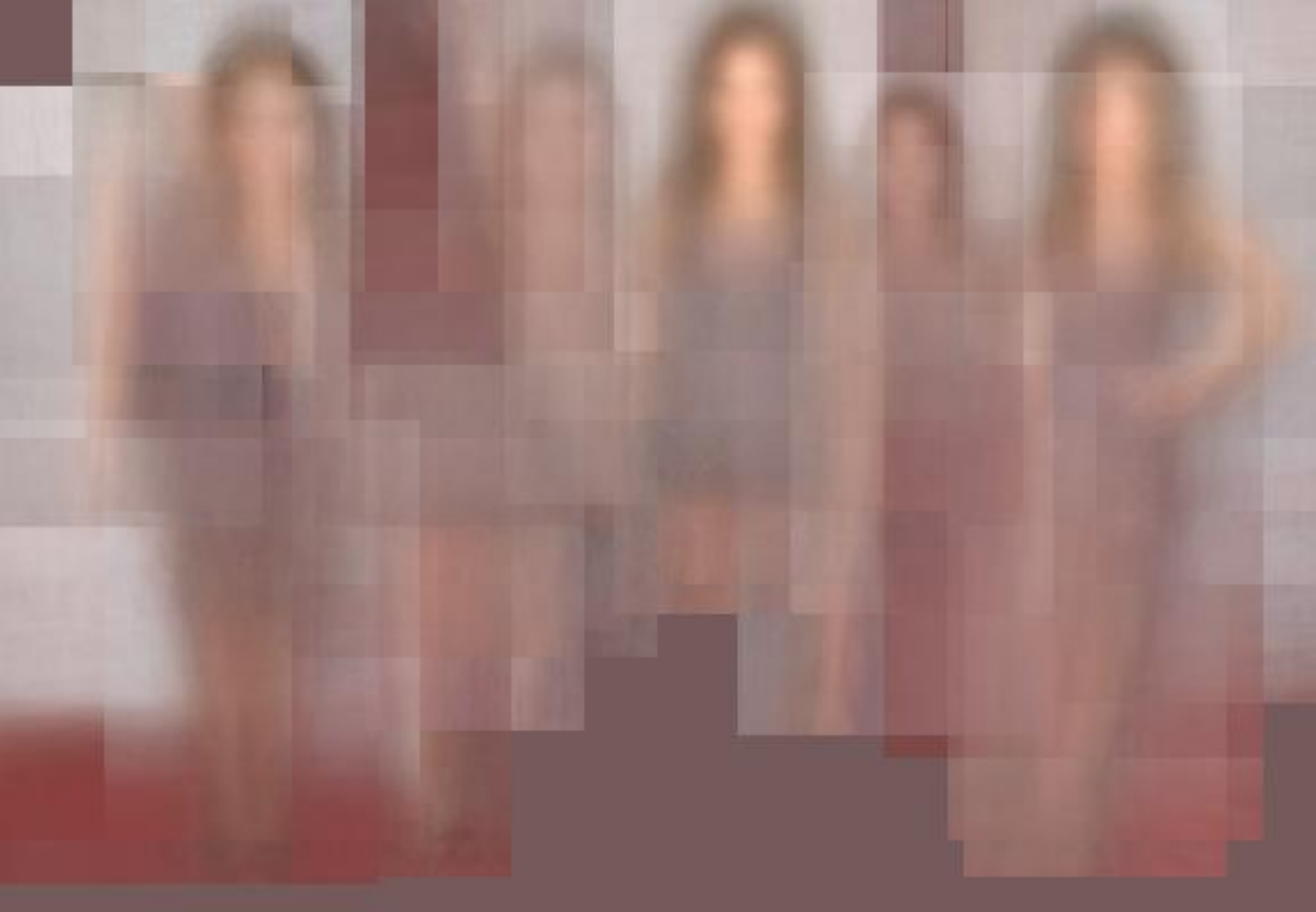}
%	}
	\hfil
%	\subfloat[Detected human viewlets from our SUVM.]{
%\subfloat[]{
%		\includegraphics[width=1.31in]{img/fullbody/1522888037783143532}%
				\includegraphics[width=1.31in]{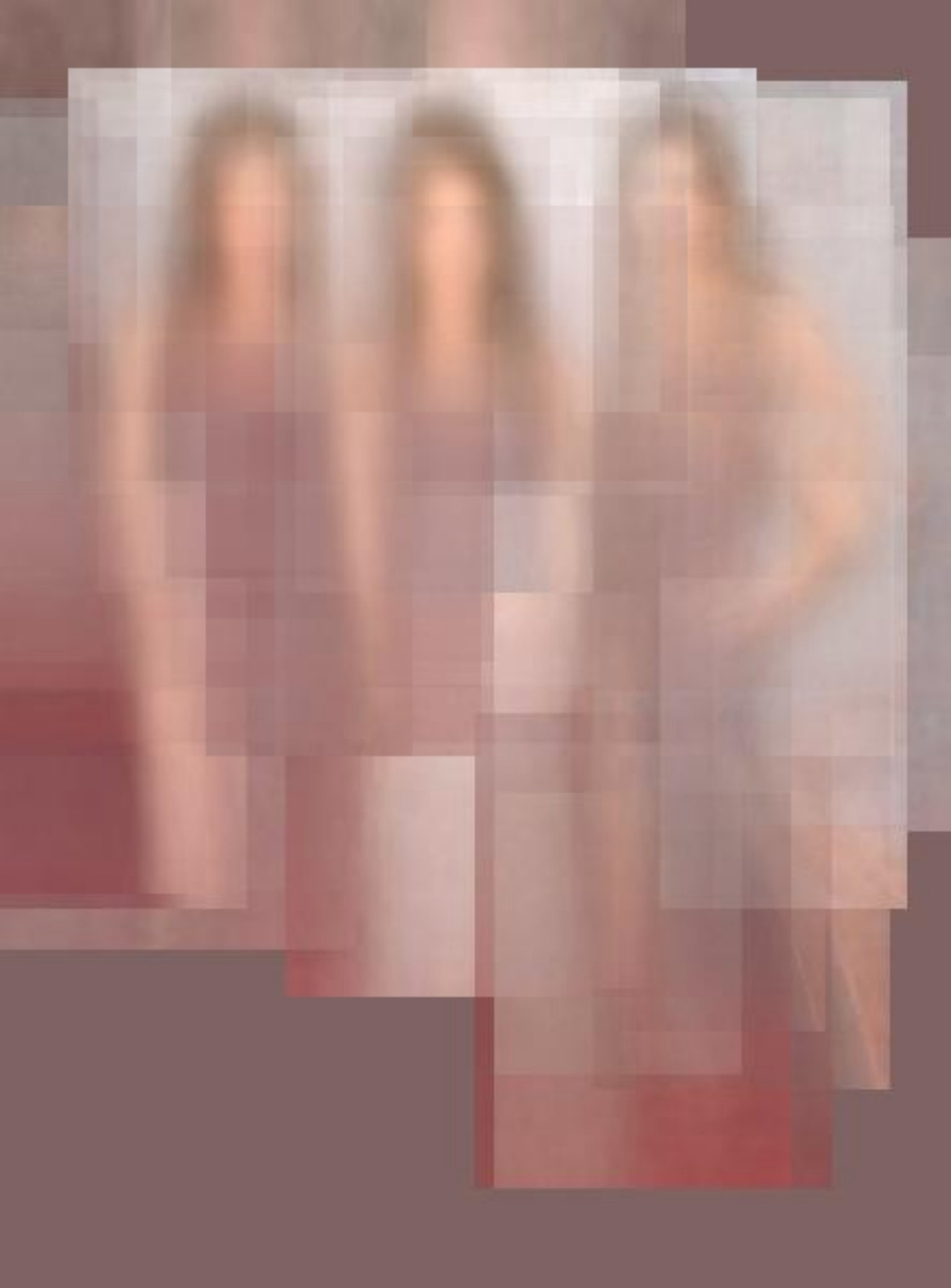}%
%	}
	\caption{An illustration of what the human SUVM sees in images with multiple people. For each individual, matching viewlets corresponding to different body parts and their poses are detected even in the presence of occlusion.}
	\label{fig:results:viewlets}
\end{figure}
\begin{figure}
	\centering
		\includegraphics[width=2.51in]{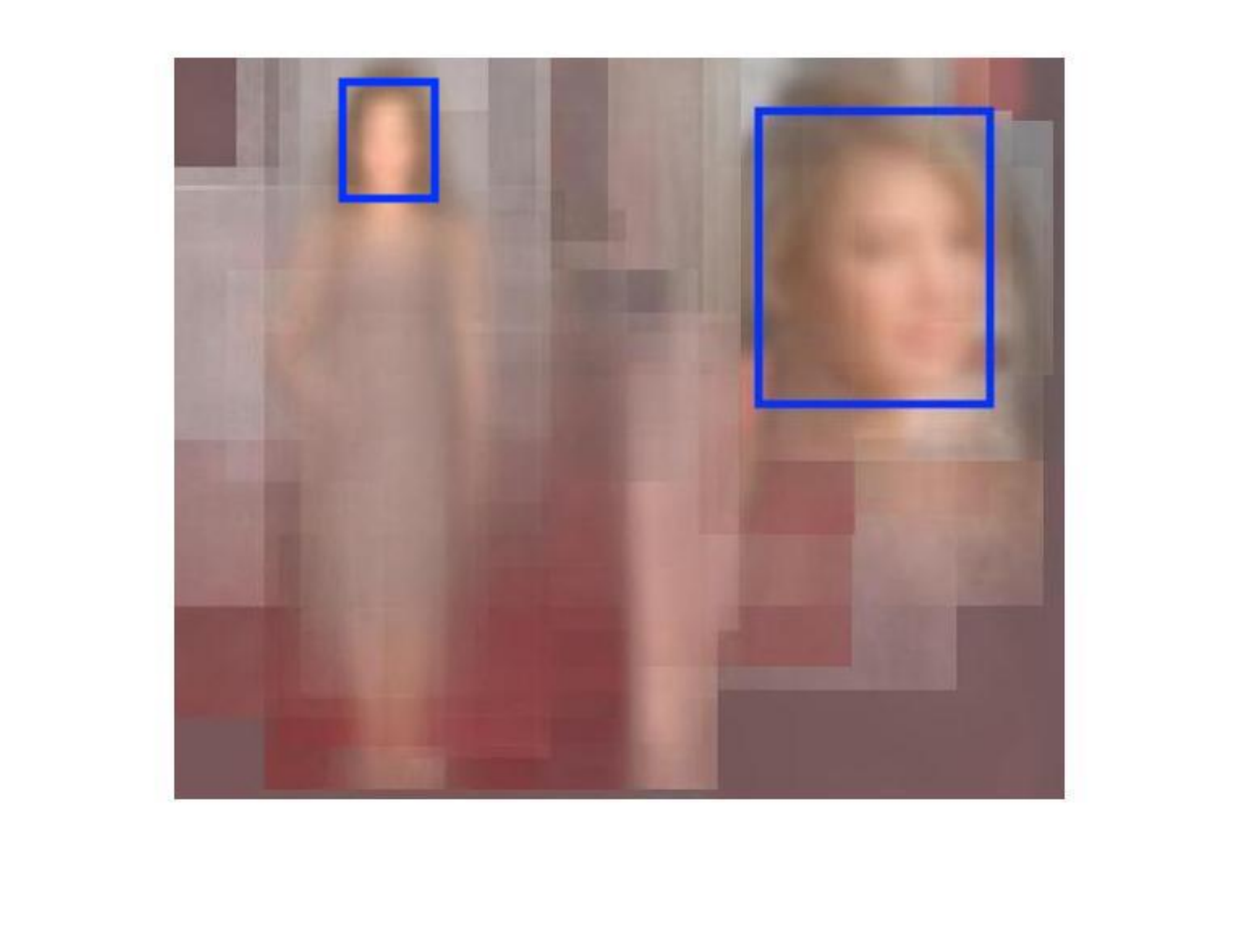}%
	\caption{Face detection at different scales and with different views (front vs side).}
	\label{fig:face:res_head_dualcele}
\end{figure}

%------------------------------------------------------------------------
\section{Discussion}
\iffalse
We have described a brain-inspired framework with several desirable properties such as (i) \textit{unsupervised} object learning and discovery capability from perceptual data (ii) \textit{computational scalability} and simplicity, and (iii) \textit{robustness} to transformations such as scaling, occlusion, and scenarios such as different views of the same three-dimensional objects. We now \iffalse highlight some of the comparative performance statistics  of the SUVM framework, and then \fi outline several follow up steps that need to be taken to extract the full power of this approach:
\fi 

\textbf{(i) Improving SUVMs:} Several features can be added to SUVMs, so as to  increase both their accuracy and power of representation. For example, currently we have not explicitly assigned prior probabilities to the occurrences of individual viewlets and parts. \iffalse (i.e., their likelihood of being detected).\fi Doing this would make detection more robust: an occurrence of a distinctive viewlet or a part could be given more weight in making a decision during object detection. Similarly, the assumption that pairwise relative distances amongst viewlets is unimodal in distribution can be relaxed. For many highly-flexible objects such as humans, the separation distance could be bimodal: for example, if we want to capture both standing and sitting postures of humans via a single SUVM, then the relative distance between head and feet viewlets would clearly have to be bimodal in distribution. Furthermore, instead of relying on random sampling and then clustering the samples, the compilation of a visual dictionary and determination of viewlets can be improved, especially using joint-segmentation and unsupervised object discovery techniques introduced for example in \cite{Cho2015},  \iffalse Furthermore, during the learning stage images could be first segmented  (in an unsupervised manner) \iffalse \cite{Arbelaez2011a})\fi into separate regions; each such resulting region could then be sampled independently to learn visual words and SUVM. This would further mimic the human object modeling process, and constrain the learned SUVMs so that they represent distinct object categories more accurately, especially when the perceptual data has multiple instances of exemplars in the same image. Our estimation and detection algorithms can be modified to efficiently account for such model enhancements and we leave that as part of our future work.
 
\hspace*{0.5cm} Another important consideration is improving the feature set used to encode viewlets. While we used the combination of HOG (Histogram of Oriented Gradients) features and k-Means clustering, in our future work we plan to explore the framework of auto-encoders used in deep learning to directly cluster the patches into visual words. 
\fi 

\hspace*{0.5cm} We also recognize that complex objects can have such radically different views so as to make them look like two different objects altogether, where they share none or very few viewlets. A sideways profile  view versus a full-body frontal view  of humans provides such an example. In such situations one can have two different SUVMs representing the two scenarios and recognize them to be the same object category based on cues other than just images. For example, spatio-temporal continuity or flow of objects in a video, where a person moves from facing the camera to facing perpendicular to it in consecutive frames, would be sufficient to determine that the views belong to the same object category. Using motion to persistently detect objects is a field of considerable interest and can be incorporated into our framework\cite{Ullman2012b}. Shared text tags in image databases can also provide such information\cite{Kang2012}.

\textbf{(ii) Dealing with low resolution images and Integrating Supervised Learning:} \label{subsec:suplearn} Traditionally a lot of the curated and publicly-available databases have predominantly low resolution images. For example, a distant silhouette of someone walking, or multiple people in the same image, where only full body views have enough resolution for detection. Detecting objects in such databases is challenging for any parts-aware approach, and especially so for ours where we use weak classifiers: there is not enough resolution to reliably detect individual viewlets corresponding to different parts, thereby losing the advantage afforded by collective decision making by a group of viewlets. For dealing with such situations, one can actively create low resolution templates from the high-resolution learned object prototypes. \iffalse As already mentioned, this approach is behind the success of the supervised Poselets methodology.\fi That is, once an SUVM is learned from high-resolution data sets, one can methodically subsample the viewlets and embed them in different scenes and build an ensemble of  classifiers using the full power of discriminative learning afforded by sophisticated classifiers, such as DNNs. The discriminative  power can be further enhanced by including  non-object visual words and image patches corresponding to viewlets belonging to other categories as negative examples. The success of a  recent approach for detecting tiny faces based on an ensemble of supervised classifiers (each classifier trained for a particular scale)\cite{Hu2017} is a good indicator  that our suggested approach would succeed. The training sets needed for our method, however, are automatically generated. minimizing the need for supervised learning. \iffalse The discriminative  power can be further enhanced by including  non-object visual words and image patches belonging to other categories as negative examples. We leave such tasks for our future work. In the meantime, however, we  have shown that with no explicit negative training and a very limited world view, the SUVM approach  \textit{is still extremely competitive for detection tasks} in images with sufficient resolution. \iffalse in addition to creating accurate and detailed object models\fi \fi

\textbf{(iii) Scaling up to detect Multiple Categories:} Using the Internet one can find large-scale perceptual data of the kind analyzed here for almost any category of objects\cite{Kang2012}. Thus, given enough data and computational power, efficient SUVMs for most individual categories can be reliably built. The preceding discussions, however, highlight that the main challenge will lie in integrating the different SUVMs: while structural information plays a very important role, it is still necessary that viewlets belonging to different object prototypes be mapped to a common feature space -- a shared visual world view -- so that they can be reliably distinguished. To facilitate such integration, \textbf{the paradigm of deep learning} with its proven capability to simultaneously learn a thousand or more different categories  can be incorporated into our framework. The \textit{manually created training sets that are currently used to train} Deep Neural Networks (DNNs), however, can now be \textit{replaced by the automatically generated viewlets}. \textit{The network will no longer be trained to detect manually-tagged categories, but to detect automatically-generated viewlets}. Thus, one can potentially \textit{automate both the tasks of object discovery and high-accuracy detection}. \iffalse The important \textit{first step is still to automatically create accurate and flexible prototypes for objects}, which is the main goal of this paper.\fi  Recent work on domain adaptation methods \cite{Gong2017} provide a proof-point for the validity of such an approach. Instead of generating an object model from a set of contextual unlabeled images (as in our work), these approaches attempt to model the overall bias between two datasets belonging to different domains. By capturing the differences between a source domain (where the classifier was trained) and the target domain (where it is applied), the performance of  updated classifiers can be improved. Here, the target domain can be unlabeled, but the source domain must be labeled and hence supervised. Our unsupervised SUVM framework can benefit from such an approach: The SUVMs for a category can be further tuned and differentiated from SUVMs for a different category in an unsupervised manner. 

\textbf{(iv) SUVMs and Recent work on Unsupervised Deep Learning (DL):} The unsupervised DL literature can be broadly categorized into three  groups: (i) \textit{Autoencoders}\cite{Hinton2007}: These generate low-dimensional representations of input signals, which can then be clustered to derive visual words in the dataset. We currently perform this step using non-DL methods: By using predefined features, such as HOG, and then by performing PCA we obtain low-dimensional feature vectors for our image patches; then we cluster these feature vectors using k-Means to obtain visual words. \textit{Our main contribution lies in creating object models that build on these visual words}, a step currently not done by the DL methods. In our ongoing work we are implementing deep autoencoders, which can provide better features and hence a more robust set of visual words.   (ii) \textit{Generative Adversarial Networks (GANs)}\cite{Goodfellow2014}: Here a DNN, driven by random noise, generates sample images that try to mimic a given set of images in an adversarial setting. If everything converges (it tends to get stuck in local minima often), then the adversarial network learns to generate outputs/images similar to those in the learning set. 
It, however, does not generate and is not intended to generate a parts-based persistent model of an object category of the type we do. (iii) \textit{Sequence Prediction:} In such a setup, given a sequence of images (for example in a video setup), one can learn to predict future frames based on current and a few past frames. Thus, the prediction network (such as the Long-Short-Term-Memory (LSTM) model\cite{Donahue2015}) can be said to have learned a representation of how objects move and change shape. This is again a useful end-to-end model that learns an over-all representation, and is not intended to learn parts-aware  representational models. In summary, the paradigm of DNNs with its ability to memorize patterns and templates is a powerful tool and in our future work we plan to use its power to make our models more accurate and expressive.  

%\iffalse 
\textbf{(v) From modeling object categories to modeling scenes} (see \cite{Geman2015} for an insightful discussion on the importance of this problem): 
%Another extension of the proposed work is to \textit{model contexts and scenes} involving multiple object categories; 
Once individual object prototypes are learned, one can go to a higher level of abstraction: instead of viewlets and their relative positions in an SUVM, one can capture the co-occurrence and relative locations and orientations of object instances (belonging to different categories) to define an analogous scene model. \iffalse A kitchen scene is therefore a loosely structured arrangement of object categories such as stoves, refrigerators, counters, dish racks etc. Just as humans do, such higher level abstractions in turn can enhance object detection itself:  any ambiguities about an individual object category can be often resolved by taking into account the context defined by the ensemble of objects in the image. A spherical object located on a stove top is most likely a designer kettle, and not a decoration piece, as it would most likely be in the context of a living room.  Again the enabling factor for modeling such scenes would be to have SUVM-type flexible models and access to large-scale visual data that cover different scenes. 
\fi 

\section{Methodology} 
\label{sec:methodology}
%------------------------------------------------------------------------
\subsection{The Structural Unsupervised Viewlets (SUV) Model (SUVM): Representation and Learning}
\label{subsec:methodology:structure}
We outline the mathematical and computational formulations of the SUVM (most of the details are deferred to the SI Appendix).

(i) \textbf{Viewlets:}  \iffalse  There is strong evidence of the presence of neurons (e.g., those in the Inferotemporal (IT) cortex) that fire selectively to  views of different parts of objects; . \iffalse Electrophysiological findings show that a particular portion of the brain (Inferotemporal cortex, or IT) appears to meet all the machinery requirement for the formation of part-based object representations.  Neurons in the Inferotemporal (IT) cortex respond selectively to stimulus from color, texture, simple structural primitives, to complex views, or even completed objects like faces\cite{Desimone1979,Mikami1980,Logothetis1996}.\fi It is, as if, the brain segments up an object into visually distinct, but potentially overlapping, jigsaw pieces of different sizes. Each such view is a building block in our model, and \iffalse , e.g, frontal vs back views of the head, or arms in different poses, or a half body view, or legs covered to different lengths.\fi  to emphasize that such views are not necessarily distinct functional parts, we refer to them as viewlets.\fi Recall that viewlets are multi-scale characteristic appearances of views of objects. To account for variations, each viewlet $V_i$ is modeled as a random variable that outputs a patch or a part of an image with  an appearance feature vector random variable $A_i$ which is drawn from a certain distribution over a feature space. \iffalse e.g. a Gaussian $\mathcal{N} (\mu_f,\Sigma_f)$ in $\mathcal{R}^{|f|}$ where $|f|$ is the dimension of the feature space.\fi For example, in this paper each sample of a viewlet $V_i$ is represented by a rectangular patch of fixed width ($w$), and fixed height ($h$).  Moreover, because viewlets can represent larger or smaller sections of the same object category under consideration (e.g., a half-body viewlet will contain the head, and hence has a larger size or scale than a head-only viewlet), we associate a relative scale parameter $S_i$ with $V_i$. To accommodate variations, $S_i$ is itself a random variable. The best way to visualize $S_i$ is to imagine a global scale parameter $s$ for an exemplar embedded in a given image, i.e., $s$ determines the overall size of the object in pixels. In such a scenario, any sample of viewlet $V_i$ has a width of  $s_i^{(x)}=w*S_i*s$ pixels and a height of $s_i^{(y)}=h*S_i*s$ pixels. \iffalse In this paper, for example, we represent viewlets as a rectangular patch of fixed dimensions, where the associated scale parameter allows for larger-size viewlets to fit in the same rectangular patch.\fi The appearance feature vector, $A_i$, of a sample is a set of  features derived from the underlying image patch.  Though for our experimental results we use local HOG (Histogram of Oriented Gradients) features (see Sections~\ref{sec:results} and SI Appendix),  \textit{ SUVMs can use any feature set}, including those derived using DNNs.

(ii) \textbf{The Spatial Relationship Network (SRN):}
Let's recall that the SRN uses a variation of the \textit{spring network model}  to represent pairwise distance and scale variations among the viewlets. Nodes in the network are the viewlets, and edges are the relative distance and scale/size constraints. To represent distances, each sample of a viewlet, $V_i$, is assigned a location coordinate, $X_i=(x_i, y_i) $, where $(x_i, y_i)$ are the pixel values of the top-left corner of the associated rectangular patch that has a width of  $s_i^{(x)}=w*S_i*s$ pixels and a height of $s_i^{(y)}=h*S_i*s$ pixels. \iffalse To be more specific, \iffalse to model the spatial relationships among the viewlets in a distributed and translation invariant manner,\fi we characterize the relative difference in locations of viewlet  pairs, i.e., we model $(X_i - X_j)$ for all pairs of viewlets $V_i$ and $V_j$, where $X_i$ and $X_j$ are the respective location parameters.\fi The relative distance between two viewlets $V_i$ and $V_j$ can then be modeled by the random variable $(X_i-X_j)$. However, in order to have \textit{both scale and translation invariance}, we need to normalize the location differences appropriately.  In particular, as explained in the SI Appendix, since the variances in the perceived/measured location coordinates (i.e., $x_i$ and $y_i$) depends on the actual lengths, we define a scale-normalized relative distance measure, $\left(\displaystyle \frac{x_i-x_j}{s_i^{(x)}+s_j^{(x)}}, \frac{y_i-y_j}{s_i^{(y)}+s_j^{(y)}}\right)$.

\iffalse we observe that $(X_i - X_j)$ is not only a function of the global scale parameter $s$ (larger the image, larger is the difference in locations) but also a function of their relative size/scale parameters $S_i$ and $S_j$. This arises due to the fact that  viewlets with larger size (as measured in pixels) will have a larger variance in its location determination, and the actual size $s_i^{(x)}=w*S_i*s$ is a function of both the global parameter $s$ and the relative scale parameter $S_i$.  As a result, to have scale invariance for each pair, we use the scale of both viewlets to do the normalization, i.e., using $\left(\displaystyle \frac{x_i-x_j}{s_i^{(x)}+s_j^{(x)}}, \frac{y_i-y_j}{s_i^{(y)}+s_j^{(y)}}\right)$ as the metric for a scale and translation invariant distance measure.\fi  

To model variations in relative positions, each pair of viewlet nodes $V_i$ and $V_j$ is connected via a spring of stiffness parameter $c_{ij}\geq 0$, and of zero-stress normalized length $\mu_{ij}$. Variations in pair-wise relative distances from their respective zero-stress lengths lead to overall stress, which can then be modeled by a potential function defined over the ensemble of springs. Further, assuming an isotropic spring model, where the displacements along  the X-axis and the Y-axis are treated separately and independently, a total potential function of a given configuration  can be written as  $\mathbf{G}=G(\mathbf{X})+G(\mathbf{Y})$, where
\vspace*{-1.5ex}
\begin{eqnarray}
\label{eq:model:model_x}
\mathbf{G}(\mathbf{X})=\frac{1}{Z^{(x)}} \exp{\left( - \frac{1}{2}\sum_{i\neq j}{c_{ij}^{(x)} \left(\frac{x_i-x_j}{s_i^{(x)}+s_j^{(x)}}-\mu_{ij}^{(x)}\right)^2} \right)},\\ %\nonumber
\label{eq:model:model_y}
\mathbf{G}(\mathbf{Y})=\frac{1}{Z^{(y)}} \exp{\left( -\frac{1}{2}\sum_{i\neq j}{c_{ij}^{(y)}\left(\frac{y_i-y_j}{s_i^{(y)}+s_j^{(y)}}-\mu^{(y)}_{ij}\right)^2}\right)},
\end{eqnarray}
%\vspace*{-1.5ex}
and (i) $Z^{(x)}$ and $Z^{(y)}$ are the corresponding normalization terms, also referred to as the partition functions, (ii)  $ \mu_{ij}^{(x)} = \left(\frac{\mu_i^{(x)}-\mu_j^{(x)}}{s_i^{(x)}+s_j^{(x)}}\right)$, $ \mu_{ij}^{(y)} =  \left(\frac{\mu_i^{(y)}-\mu_j^{(x)}}{s_i^{(y)}+s_j^{(y)}}\right)$ are the normalized zero-stress lengths. 
The above expressions are functions of the relative scale parameters $S_i$, and their pair-wise variations can again be modeled via springs. Since scale is  a multiplicative factor, we take its logarithm and define an analogous potential function,
\vspace*{-1.5ex}
\begin{eqnarray}
\label{eq:model:model_s}
G(\mathbf{S})=\frac{1}{Z^{(s)}} \exp{\left( -\frac{1}{2}\sum_{i\neq j}{c^{(s)}_{ij}\left(\log{\frac{S_i}{S_j}}-\log{\frac{\mu_i^{(s)}}{\mu_j^{(s)}}}\right)^2} \right)} ,
\end{eqnarray} 
%\vspace*{-1.5ex}
where $\mu_i^{(s)}$ and $\mu_j^{(s)}$ are the respective expected scales of viewlets $V_i$ and $V_j$.

\textbf{Gaussian Markov Random Field Model (GMRF):} \label{subsec:torso-gmrf} A Markov Random Field is a set of random variables having a Markov property described by an undirected graph. In particular, each random variable node is independent of the rest of the random variables given its neighbors in the network. If the joint distribution is Gaussian, with joint covariance matrix, $\Sigma$, then the GMRF specifies the zero patterns of the  \textit{Precision matrix} $\Lambda = \Sigma^{-1}$:  $\Lambda_{ij}=0$ implies that the corresponding random variables are conditionally independent and hence does not have an edge in the GMRF. For quadratic-form potential functions (as in the above equations), we can regard our spring model as a \textit{Gaussian Markov random field}. \iffalse , where instead of specifying the joint Covariance matrix, $\Sigma$, \textit{the network model directly specifies the Precision matrix} $\Lambda = \Sigma^{-1}$ via the edges of the network. We start with the $X$-axis displacement potential function, and the same will hold for the $Y$ and $S$ potential functions.\fi The precision matrix $\Lambda^{(x)}$ can be calculated by noting that $\Lambda_{ij}^{(x)}$ is the coefficient of the product terms $(x_i-\mu_i^{(x)})(x_j-\mu_j^{(x)})$ in the exponent of  Equation~\ref{eq:model:model_x}: 
\vspace*{-1.5ex}
\begin{eqnarray}
\label{eq:Lambda_x}
\Lambda_{ii}^{(x)} = \sum_{j=1,j\neq i}^{M-1}\frac{c_{ij}^{(x)}}{(s_i^{(x)}+s_j^{(x)})^2}+\frac{c_{iM}^{(x)}}{(s_i^{(x)}+s_M^{(x)})^2},\\ \Lambda_{ij}^{(x)} = -\frac{c_{ij}^{(x)}}{(s_i^{(x)}+s_j^{(x)})^2} \quad i \neq j\ ,
\end{eqnarray} 
where without loss of generality, we have assumed $X_M=0$ to reduce the degree of freedom to $M-1$ (see SI Appendix). \iffalse Since the precision matrix is an $M$-matrix (i.e., all the off-diagonal entries are non-positive and it is diagonally dominant), the corresponding covariance matrix is constrained to be non-negative.\fi Note that if $c_{ij}=0$ then $\Lambda_{ij} =0$ and we know from the properties of multi-variate Gaussian distributions that the corresponding location variables are conditionally independent. Now Eq.~\ref{eq:model:model_x} can be written as a log-likelihood function:
\vspace*{-1.5ex}
\begin{eqnarray}
\label{eq:model:ll_x} {\cal{L}}(X)= \frac{1}{2} \log{r|\Lambda|}-\frac{1}{2}\sum_{i\neq j}{c_{ij}^{(x)} \left(\frac{x_i-x_j}{s_i^{(x)}+s_j^{(x)}}-\mu_{ij}^{(x)}\right)^2}, 
\end{eqnarray}
where $|\Lambda|$ is the determinant of the precision matrix, and $r$ is the normalization constant for a Gaussian distribution. \\
\textbf{Sparsity and Conditional Independence:} 
The direct interactions (i.e., for which $c_{ij}> 0$), combined with node set, $V$, form the SRN network, $G(V,E)$, and hence the model complexity of the SRN corresponds to its sparsity.  Sparsity has a physical meaning in our model: For most physical objects, locations of parts and the resulting views are indeed not statistically fully connected with each other. Equivalently, a sparse set of springs are enough to constrain deformations in exemplars. \iffalse We, thus, expect that most objects can be modeled by $\Theta(n)$ edges where $n$ is the number of viewlets.\fi We impose such sparsity constraints in the learning process, and a relaxation leads to a convex optimization problem that can be solved efficiently. \iffalse We, however show that  perceptual quantities such as variances in the relative locations of viewlets, provide efficient heuristic ways of approximating solutions to the convex optimization problem. \fi 

(iii) \textbf{Semantic Structure:} As already explained in the introduction, 
we use two complementary constructs to capture the semantic structure of the object, namely the  \textit{Configurable-Independent Parts Clustering} (CIPC), and the \textit{Global Positional Embedding (GPE)}. The specifics of these two constructs are made precise in the learning section. For now, it suffices to mention that together they provide a description of the object prototype in terms parts (each part being a grouping of viewlets), their locations, and the inclusion/overlap relationships among the viewlets and parts.

(iv) \textbf{\iffalse A Probabilistic Interpretation\fi Generative model and Calculating Object Likelihoods:} An SUVM defined by its parameter set $\theta = \left(\{c_{ij}^{(x)}\},\{c_{ij}^{(y)}\},\{c_{ij}^{(s)}\}, \{\mu_{i}^{(x)}\}, \{\mu_{i}^{(y)}\}, \{\mu_{i}^{(s)}\}\right)$ that specifies the SRN \iffalse(let $n_f$ be the number of viewlets in the SRN)\fi and the accompanying CIPC and GPE models,  is the key representational and generative tool we  use. \iffalse for abstracting how brains model and view real world objects.\fi  As a generative model, any exemplar can be  viewed as being created by a four-step process: (i) First picking the parts or regions that are to be rendered in the exemplar from the CIPC and GPE; let $D_p$ be the set of parts that is picked; (ii) Then picking $N_G$ viewlets, numbered $1, \cdots, N_G$,  that go together for the picked parts; for example, for configurable parts certain viewlets are mutually exclusive and should not be picked together. Let $V_G$ be the set of picked viewlets. The probability $P(V_G|\theta)$ is stated in the SI Appendix, where we provide a detailed description of our detection algorithms. For each picked viewlet, $V_i\in V_G$, an appearance feature vector $A_i$ is drawn by sampling its appearance distribution, and a corresponding image patch is created; let $A=\left\{A_1, \cdots, A_{N_G}\right\}$. (iii) Then choosing scaling factors by sampling the joint scale distribution (Eq.~\ref{eq:model:model_s}); let $S=\left\{S_1, \cdots, S_{N_G}\right\}$,  and finally (iv) locating these $N$ viewlets spatially by sampling the joint distribution specified by the SRN (Eqs.~\ref{eq:model:model_x} and \ref{eq:model:model_y}). Let $X=\left\{x_1, \cdots, x_{N_G}\right\}$ and $Y=\left\{y_1, \cdots, y_{N_G}\right\}$ be the set of these location coordinates. Note that in our model, given $S$ and $V_G$, $X$ and $Y$ are picked independently. \\ Each step has its own likelihood allowing us to calculate the likelihood of any such generated exemplar:
%\vspace*{-1.25ex}
\begin{eqnarray}
P(\hbox{Generated Exemplar}|\theta) = P(A,X,Y,S,V_G|\theta) = \nonumber \\
P(Y|S,V_G, \theta) \times P(X|S,V_G, \theta)\nonumber \\ \times P(A|V_G,\theta)\times P(S|V_G,\theta) \times P(V_G|\theta)\ .
\label{eq:detect:mo_fore}
\end{eqnarray}
%------------------------------------------------------------------------
\subsection{Learning SUVMs}
\label{subsec:methodology:learn}
Given a learning dataset comprising unlabeled instances of the unknown category, we need to construct an SUVM, i.e., appearance feature vectors of the viewlets, the SRN and the semantic structures, CIPC and GPE.

(i) \textbf{Learning A Visual Dictionary:} In this step we determine a set of visual words, or a dictionary, from the given images. We randomly sample all images in the learning set utilizing a scale pyramid and using a fixed-size rectangular patch, then convert all patches into image feature vectors, and then extract a visual vocabulary out of them using an unsupervised clustering algorithm.  Note that each visual word is a cluster of feature vectors, and the visual dictionary naturally comes with a classification algorithm. For example, for k-Means clustering one can use the k-Nearest-Neighbor (kNN) algorithm to assign a word label to any candidate image patch. Also note that a visual word  represents only a \textit{potential viewlet} in the object models to be extracted from the learning set. 

(ii) \textbf{A Maximum Likelihood (ML) framework for Learning SRNs (Spatial Relation Networks): }
We have already simplified our model in the form of a sparse network, or a GMRF, which can be determined by an edge set $E$, and the related parameters $\left(\{c_{ij}^{(x)}\},\{c_{ij}^{(y)}\},\{c_{ij}^{(s)}\}, \{\mu_{i}^{(x)}\}, \{\mu_{i}^{(y)}\}, \{\mu_{i}^{(s)}\}\right)$. Since we have already created a set of visual words, we first go back to the original image corpus (e.g., in the celebrity data set, {9638}  images are in  the learning set, see Section~\ref{sec:results}) and detect in each image the visual words that appear in it. That is, in every image, we first perform a dense scan (using a scaling pyramid so that we  capture viewlets that have inherently larger scale),  with a fixed-size sliding window (the same size as used to determine the visual dictionary), and assign a visual word to each resulting patch using a k-Nearest-Neighbor (kNN) algorithm. Then, for each detected visual word, $V_i$, we have its size in pixels ($s_i^{(x)}. s_i^{(y)}$), and its location coordinates $(x_i, y_i)$.  For a pair of visual words, $V_i$ and $V_j$, detected in the same image,  we have samples of the SUV model outputs: $Z^{(s)}_{ij}  = \frac{S_j}{S_i}= \frac{s^{(x)}_j}{s^{(x)}_i}={\frac{s^{(y)}_j}{s^{(y)}_i}}$;  $Z^{(x)}_{ij} = \frac{(x_j-x_i)}{(s_i^{(x)} + s_j^{(x)})}$; and 
$Z^{(y)}_{ij} = \frac{(y_j-y_i)}{(s_i^{(y)} + s_j^{(y)})}$. We need to infer now an edge set $E$, and the related spring parameters such that the data likelihood, as captured by equations~\ref{eq:model:model_x}--\ref{eq:model:model_s}, is maximized. For example, to estimate the set of parameters $\{c_{ij}^{(x)}\}$, it follows from preceding discussions on GMRF on Page ~\pageref{subsec:torso-gmrf} that the empirical likelihood function is given by:  $\log{P(X)}= \mbox{const} + \frac{1}{2} \log{|\Lambda|}-\frac{1}{2}\sum_{i\ne j}{c_{ij}^{(x)}\operatorname{Var}(Z_{ij})}$, where $\operatorname{Var}(Z_{ij})$ is the empirically observed variance of the random variable $Z_{ij}$.

\textbf{Approximate Sparse Estimation of $c_{ij}$'s: \iffalse using $L_1$ regularization and Convex Relaxation\fi} To maximize $\log{P(X)}$ while making $c_{ij}$'s sparse, we reverse the sign to get a minimization problem and add an $L_1$ regularization term to obtain: $ {\cal{L}}(X)= -\frac{1}{2}\log{|\Lambda|}+\frac{1}{2}\sum_{i\ne j}{c_{ij}(\operatorname{Var}(Z_{ij})+\lambda)},$
where $\lambda>0$ is the regularization parameter and $c_{ij} \geq 0$.  \textit{This is a convex optimization problem} and can be solved efficiently. Staying true to our spirit of performing simple computations, we analyze this convex optimization problem and using the KKT conditions, we prove an upper bound on the optimal values of $c_{ij}$: $c_{ij}^* \leq \frac{1}{\operatorname{Var}(Z_{ij})+\lambda}$ (see SI Appendix). Thus,  $c_{ij}^*$ decreases monotonically with increases in both the observed variance, $\operatorname{Var}(Z_{ij})$, and the sparsity parameter, $\lambda$. \iffalse This makes intuitive sense, as two different viewlets that correspond to parts that are not directly linked by a stiff joint or some such structure in the physical object  (hence, there are intermediate parts connecting them, and the location of one could be predicted accurately, given the locations of these intermediate parts) will tend to have a higher variance in their relative locations.\fi  This bound then leads to an approximate but efficient algorithm to directly impose sparsity: if we say that all those edges for which the optimal $c_{ij}^*$ is  less than say a target value of $c$ will be removed from the network, then it implies from the above equation that all edges with empirical $\operatorname{Var}(Z_{ij}) >\frac{1}{{c}}-\lambda$ should be disconnected, or their corresponding $c_{ij}=0$. Thus, we have derived a simple threshold rule on the pairwise variances, and by lowering the threshold (that is, by increasing the sparsity parameter $\lambda$) we get increasingly sparse SRN's. In our implementation, we defined a combined variance for an edge $V = \operatorname{Var}({Z}^{x}_{ij})+\operatorname{Var}({Z}^{(y)}_{ij}) + \operatorname{Var}(\log Z^{(s)}_{ij})$ and then imposed a threshold on it until the the SRN is sparse enough. Note that as the edge set becomes sparse, the initial network, comprising all visual words, gets disconnected and the giant connected components  correspond to the SRNs of the underlying object categories. 

%------------------------------------------------------------------------
(iii) \textbf{Extracting Parts Using {Configuration-Independent Parts Clustering} (CIPC):}
%\label{subsec:model:learn_sen}
We first point out two ways in which a ``part'' in the object category gets encoded in terms of viewlets and their structure in our model:  (i) Two or more viewlets that are replaceable in making up the whole object, or equivalently, two or more viewlets that are \textit{mutually exclusive} (in terms of co-occurrence and hence shares no edge in the SRN), \textbf{and yet} have \textit{nearly-identical geometrical relationships} with other viewlets (representing other parts). Pairs of such nodes/viewlets  can be identified efficiently by processing the SRN, e.g. by sequentially examining each viewlet node and finding other nodes that are not connected to it by an edge but share neighbors in common;   (ii)  \iffalse In order to make the SEN robust we also add a different group of edges that can be considered as dual to the other set of edges: This scenario arises when two viewlet nodes are only slightly shifted versions of each other. \fi Viewlets that share a very stable edge in the SRN between them, and have the same geometrical relationships with viewlets corresponding to other parts of the object. This scenario arises when two viewlet nodes are only slightly shifted versions of each other, representing persistent presence of a part in the object. Again such pairs can be efficiently detected from the SRN. We construct a \textit{CIPC network}, where each pair of viewlets, satisfying type (i) or (ii) relationship, are connected by an edge. \textit{Each  connected component in the resulting CIPC network  then corresponds to a distinct configurable and stable part of the underlying object category}. The results shown in Figs~\ref{fig:introduction:gps} and in the SI Appendix demonstrate the effectiveness of this methodology. 

(iv) \textbf{Extracting Semantic Structure Using GPE (Global Positional Embedding):}
%\label{subsec:model:learn_gps}
In this step we use the pairwise scale and location relationships to embed the viewlets in the SRN in a 3D space: Each viewlet $V_i$ is assigned an absolute 2-D positing $(x(i), y(i))$ and scale $S(i)$. such that the pairwise constraints obtained from data are best satisfied. We use a mean-squared error based optimization function,  similar to that used in Multi-Dimensional Scaling (MDS)\cite{Kruskal1964} and derive an iterative approach to calculate these mean positions and scales of viewlets (See SI Appendix). \iffalse using the relative positions and sizes of its neighbors (on the SRN) by solving a mean-squared error based optimization problem \fi  

From Figure~\ref{fig:introduction:gps} we notice that viewlets, clustered by CIPC as belonging to the same part, have very similar global spatial values and cluster together in the GPE. \iffalse Furthermore, the global positions of the viewlets in these communities of the CIPC network mirrored the real world object structures.\fi Our ability to reverse engineer human body parts, for example, demonstrates that we are able to identify the semantic structure of objects automatically, instead of hand-coding such knowledge via manual tagging. 

%------------------------------------------------------------------------
\vspace*{-1ex}
\subsection{From Models To Detection}
\label{subsec:methodology:detect}
We start with a dense scanning of the given image using a scale pyramid and obtain $N$ patches; note that $N$ can easily be in the thousands. Next, we design complementary algorithms for two different tasks for any given image: (i) \textbf{Task 1:} Detection and localization of  object instances (for example, cars, humans etc.), and (ii) \textbf{Task 2:} Detection and localization of  specific parts (e.g., human head/face or torso) of a learned object category. In both cases there could be multiple occurrences in the same image. For \textbf{Task 1}, following the probabilistic interpretation 
introduced in Eq.~\ref{eq:detect:mo_fore}, we do the following search: (i) we map each image patch to a visual word and consider only those that are mapped to viewlets in the SUVM, and then (ii) group the viewlet patches  into clusters so that each cluster of patches \iffalse (with  their appearance descriptors, locations, and scales)\fi maximizes the likelihood of representing an object instance, as given Eq.~\ref{eq:detect:mo_fore}. This can be accomplished via an exponential search over all possible assignments of image patches to visual words \cite{Felzenszwalb2008,Felzenszwalb2010,Fergus2006}. In accordance with our neuroscience inspirations, however, we do a restricted search and  we consider an object to be detected if sufficiently many parts (as determined by CIPC and GPE) that match the relative distances and scale requirements are detected with high confidence. Thus, we look at all the detected viewlets, and then start grouping them together based on whether they structurally match our model. This linear time (in $N$ and $n_f$) heuristic search algorithm is agglomerative in nature rather than exhaustive, making it highly scalable.   Moreover, this natural detection framework allows one to find multiple occurrences of objects in the same image efficiently. The details are given in the SI Appendix.

For \textbf{Task 2} where the goal is to detect and localize targeted parts  of learned object instances, we again use the structure of the underlying SUVM to compute a geometric mapping between any given pair of viewlets, $V_i$ and $V_j$ (see SI Appendix for details). Thus a reliable target part and its location is detected by mapping multiple detected viewlets to where the  part should be. 
\vspace*{1ex}
\section{Acknowledgement} The authors thank Prof. Lieven Vandenberghe for his input on the optimization formulations used in the paper, and also the referees for helpful suggestions and especially for pointing us to relevant prior work.

% Bibliography
%\bibliography{dissertation}

\end{document}